\documentclass[letterpaper]{article} 
\usepackage{aaai23} 
\usepackage{times}  
\usepackage{helvet}  
\usepackage{courier}  
\usepackage[hyphens]{url}  
\usepackage{graphicx} 
\usepackage{natbib}  
\usepackage{caption} 
\usepackage{algorithm}
\usepackage{algorithmic}

\usepackage{tikz}
\usepackage{comment}
\usepackage{amsmath,amssymb} 
\usepackage{booktabs}
\usepackage{cite}
\usepackage{multirow}
\usepackage[normalem]{ulem}
\useunder{\uline}{\ul}{}
\usepackage{color}
\usepackage{wrapfig}
\usepackage{subcaption}

\usepackage{newfloat}
\usepackage{listings}
\DeclareCaptionStyle{ruled}{labelfont=normalfont,labelsep=colon,strut=off} 
\floatstyle{ruled}
\newfloat{listing}{tb}{lst}{}
\floatname{listing}{Listing}
\title{Target-Free Text-guided Image Manipulation}

\author {
    Wan-Cyuan Fan\textsuperscript{\rm 1},
    Cheng-Fu Yang\textsuperscript{\rm 2}, 
    Chiao-An Yang\textsuperscript{\rm 3},
    Yu-Chiang Frank Wang\textsuperscript{\rm 1, \rm 4}
}
\affiliations {
    \textsuperscript{\rm 1}National Taiwan University \
    \textsuperscript{\rm 2}UCLA \ 
    \textsuperscript{\rm 3}Purdue University \ 
    \textsuperscript{\rm 4}NVIDIA\\
    \texttt{\small r09942092@ntu.edu.tw} \ \ \ \texttt{\small cfyang@cs.ucla.edu} \ \ \  \texttt{\small yang2300@purdue.edu} \ \ \  \texttt{\small frankwang@nvidia.com}
}

\usepackage{bibentry}

\begin{document}

\maketitle

\begin{abstract}
    We tackle the problem of target-free text-guided image manipulation, which requires one to modify the input reference image based on the given text instruction, while no ground truth target image is observed during training. To address this challenging task, we propose a Cyclic-Manipulation GAN (cManiGAN) in this paper, which is able to realize \textit{where} and \textit{how} to edit the image regions of interest. Specifically, the image editor in cManiGAN learns to identify and complete the input image, while cross-modal interpreter and reasoner are deployed to verify the semantic correctness of the output image based on the input instruction. While the former utilizes factual/counterfactual description learning for authenticating the image semantics, the latter predicts the ``undo" instruction and provides pixel-level supervision for the training of cManiGAN. With such operational cycle-consistency, our cManiGAN can be trained in the above weakly supervised setting. We conduct extensive experiments on the datasets of CLEVR and COCO, and the effectiveness and generalizability of our proposed method can be successfully verified. Project page: \textcolor{blue}{\url{sites.google.com/view/wancyuanfan/projects/cmanigan}}.

\vspace{-2mm}

\end{abstract}

\section{Introduction}
\label{sec:intro}

\begin{table*}[t]
\centering
\resizebox{\textwidth}{!}{
\begin{tabular}{c|cccc|ccc}
\hline
   \multirow{2}{*}{Methods}
   & \multicolumn{4}{c|}{Input data}
   & \multicolumn{3}{c}{Manipulation type}
   
   \\ \cline{2-8} 
      & Instruction
      & Description
      & GT image
      & Auxiliary info
      & Change visual attribute
      & Remove object
      & Add object   \\\hline
      
ManiGAN~\cite{li2020manigan} &-&$\checkmark$
&No&-&$\checkmark$&-&- \\
TediGAN~\cite{xia2021tedigan} &-&$\checkmark$
&No&-&$\checkmark$&-&- \\
\hline
ASE~\cite{shetty2018adversarial} &-&-&No&Image-level
labels&-&$\checkmark$&-\\
GeNeVa~\cite{el2019tell} &$\checkmark$&-&Yes&-
&-&-&$\checkmark$\\
TIM-GAN~\cite{zhang2021text} &$\checkmark$&-&Yes&-
&$\checkmark$&$\checkmark$&$\checkmark$\\
Ours &$\checkmark$&-&No&Image-level
labels
&$\checkmark$&$\checkmark$&$\checkmark$\\

\hline
\end{tabular}
}
\vspace{-3mm}
\caption{Comparisons of recent approaches on text-guided image manipulation. Note that GT image indicates the need of ground-truth target image during training.}
\label{tab:com_1}
\vspace{-6mm}
\end{table*}

Image manipulation by text instruction (or text-guided image manipulation) aims to edit the input reference image based on the given \textit{instruction} that describes the desirable modification to the image. This task not only benefits a variety of applications including computer-aided design~\cite{el2019tell, viazovetskyi2020stylegan2}, face generation~\cite{xia2021towards, xia2021tedigan} and image editing~\cite{zhang2021text, shetty2018adversarial, li2020manigan, wang2021cycle, patashnik2021styleclip, li2020lightweight}, the developed algorithm can be further applied as a data augmentation technique for learning deep neural networks. In addition to the need to output high-quality images, the main challenges in text-guided image manipulation are to identify ``\textit{where}" and to know ``\textit{how}" to edit the image based on the given instruction. In other words, how to bridge the gap between semantic and linguistic information during image manipulation process requires the efforts from researchers in related fields.

Existing text-to-image manipulation works can be divided into two categories: \textit{object-centric image editing} and \textit{scene-level image manipulation}~\cite{shetty2018adversarial}. For object-centric image editing~\cite{choi2018stargan, li2019controllable, li2020lightweight, li2020manigan}, existing works focus on modifying visual attributes (e.g., color or texture) of particular objects (e.g., face or bird) in the image, or to change its style (e.g., age or expression) to match the given \textit{description} (not necessarily \textit{instruction}). With such image-description pairs observed during training, the text-guided editing process can be simply achieved by mapping the information between the images and the corresponding descriptions. While ground truth target image is not necessarily required, such models require descriptions for both images before and after editing.

As for scene-level image manipulation~\cite{el2019tell, shetty2018adversarial, zhang2021text, dhamo2020semantic}, its goal is to reorganize the composition of the input image (e.g., moving, adding, and removing objects in the images). Since the input image might contain multiple objects in the scene, to localize ``where" to edit would be a difficult task to handle. Moreover, instead of changing attributes of a given object, the model needs to generate objects or introduce a background on the location of interest. Thus, Zhang et al.~\cite{zhang2021text} decompose the above manipulation process into two stages: localization and generation. With target images as supervision, TIM-GAN~\cite{zhang2021text} is trained to manipulate the reference image with visual realism and semantic correctness. However, since the ground truth target image might \textit{not} always be available, Adversarial Scene Editing (ASE)~\cite{shetty2018adversarial} chooses a weakly supervised setting with image-level labels as weak guidance. Nevertheless, ASE only allows one to \textit{remove} an object in the scene and cannot easily be applied to operations like adding or changing an attribute.

In this paper, we propose a cyclic-Manipulation GAN (cManiGAN) for target-free image manipulation. Due to the absence of ground truth target image during training, it is extremely challenging to identify \textit{where} and \textit{how} to edit the input reference image, so that the output would be semantically correct. To tackle the above two obstacles, our cManiGAN exploits self-supervised learning for enforcing the semantic correctness, while pixel-level guidance can be automatically observed. More specifically, the image editor of cManiGAN learns to locate/edit the image region of interest with global and local semantics observed. And, the modules of cross-modality interpreter and reasoner are deployed in cManiGAN. The former is introduced to verify the image semantics via factual/counter-factual description learning, while the latter infers the ``undo" instruction and offers pixel-level supervision. As detailed later, the above design uniquely utilizes cross-modal cycle consistency and allows the weakly-supervised training our of cManiGAN.

\section{Related Works}

\noindent \textbf{Object-centric Image Manipulation by Text Instruction.} ControlGAN~\cite{li2019controllable} is an end-to-end trainable network for synthesizing high-quality images, with image regions fitting the given descriptions. Li et al. propose~\cite{li2020manigan, li2020lightweight}, containing an affine combination module (ACM) and a detail correction module (DCM), which manipulate image regions based on both input text and the desired attributes (e.g., color and texture). Liu et al.~\cite{liu2020open} utilize unified visual-semantic embedding space so that manipulation can be achieved by performing text-guided vector arithmetic operations. Recently, Xia et al.~\cite{xia2021tedigan} apply StyleGAN to edit the reference image via instance-level optimization. With the given text as guidance, its produced image would be close to the reference input in the embedding space.\\

\noindent \textbf{Scene-level Image Manipulation by Text Instruction.} El-Nouby et al.~\cite{el2019tell} propose GeNeVa for the serial story image manipulation, which sequentially predicts new objects based on the associated descriptions to a story scene background. Dhamo et al. propose SIMSG~\cite{dhamo2020semantic}, which encodes image semantic information into a given scene graph for manipulation purposes. Recently, TIM-GAN~\cite{zhang2021text} decomposes the manipulation process into localization and generation. The introduced Routing-Neurons network in the generation exhibits the ability to dynamically adapt different learning blocks for different complex instructions, better capturing text information and thus with improved manipulation ability. As noted earlier, existing methods for scene-level image manipulation require reference-target training image pairs (i.e., a fully supervised setting). While methods such as ManiGAN~\cite{li2020manigan} and TediGAN~\cite{xia2021tedigan} do not observe the target images during training, these methods are generally applied to object-centric images and cannot be easily generalized to perform scene-level manipulation.

To alleviate the above concern, Shetty et al.~\cite{shetty2018adversarial} introduce ASE, allowing users to remove an object in a scene-level image while not requiring pairwise training data. Instead of producing an entire image output based on given text~\cite{li2019controllable, reed2016generative} and scene graphs~\cite{herzig2020learning, johnson2018image, yang2021layouttransformer}, ASE focuses on generating the background of a specific area on the image to smoothly remove the target object. As listed in Table~\ref{tab:com_1}, we compare the properties of recent image manipulation approaches and highlight the practicality of ours.

\begin{figure*}[t]
  \centering
  \includegraphics[page=1,trim={30 60 30 70}, clip, width=0.95\textwidth]{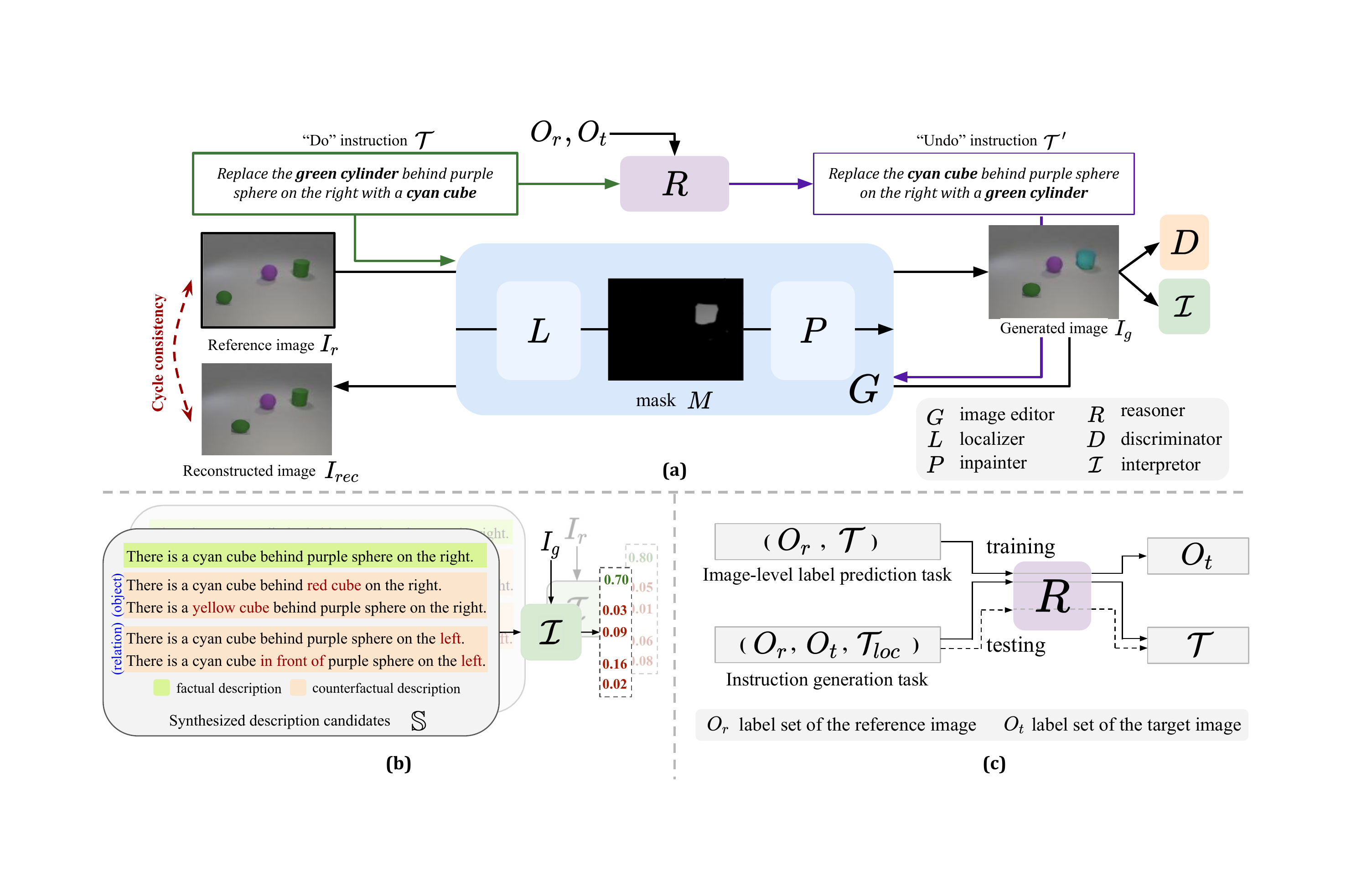}
  \vspace{-5mm}
  \caption{(a) \textbf{Architecture of cManiGAN}, which consists of generator $G$ (with localizer $L$ and inpainter $P$), discriminator $D$, cross-modal interpreter $\mathcal{I}$, and reasoner $R$. Note that $O_r$ and $O_t$ are image-level labels for the reference and target images, respectively.
  (b) \textbf{Cross-modal interpreter} $\mathcal{I}$ in (a), which authenticates the output image via factual/counterfactual descriptions. (c) \textbf{Reasoner $R$} deployed in (a) to produce the undo instruction for cross-modal cycle consistency. Note that $\mathcal{T}_{loc}$ is the adverbs of place part of instruction $\mathcal{T}$. Please see Methodology for detailed discussions for each module.}
  \vspace{-5mm}
  \label{Architecture_2}
\end{figure*}

\section{Methodology}
\subsection{Notations and Algorithmic Overview}
In this work, one is given a reference image $I_r$ and the associated instruction $\mathcal{T}$, describing where and how to edit $I_r$. \textit{Without} observing the ground truth target image $I_t$, our goal is to produce an image $I_g$ matching $\mathcal{T}$ (i.e., with the desirable layout and/or attributes). 

To tackle this problem, we propose a GAN-based framework of Cyclic Manipulation GAN (cManiGAN),  as shown in Fig.~\ref{Architecture_2}(a). Given $I_r$ and $\mathcal{T}$, our cManiGAN has an image editor (generator) $G$, which consists of a spatial-aware localizer $L$ and an image in-painter $P$. The former is deployed to identify the target object/attributes of interest by producing a binary mask $M$, and the latter is to complete the output image $I_g$ accordingly. To enforce the correctness of the visual attribute and the location of the generated object, a cross-modal \textit{interpreter} $\mathcal{I}$ is introduced in cMAniGAN, which learns to distinguish between factual and synthesized counterfactual descriptions. Moreover, an instruction \textit{reasoner} $R$ is deployed to infer the ``undo" instruction $\mathcal{T}'$ from  $\mathcal{T}$. With this ``undo" instruction, a cyclic-consistent training scheme can be conducted which provides pixel-level supervision from the recovered image output.

Following~\cite{shetty2018adversarial}, we utilize the image-level labels (i.e., $O_r$ and $O_t$) from the reference and target images as weak supervision. Note that such labels are practically easy to obtain, since they can be produced by pre-trained classifiers or by rule-based language models like~\cite{manning2014stanford} to infer the labels from $I_r$ and $\mathcal{T}$ (as discussed in the supplementary materials).

\subsection{Image Editor $G$}
An overview of the proposed Cyclic-Manipulation GAN (cManiGAN) is illustrated in Fig.~\ref{Architecture_2}(a). The generator, or image editor $G$, aims to modify the given reference image $I_r$ based on the instruction $\mathcal{T}$ and produces the manipulated result $I_g$. We now detail the design and learning objectives for the generator $G$.\\

\subsubsection{Localizer $L$.} As shown in Fig.~\ref{Architecture_2}(a), a spatial-aware localizer $L$ is deployed in the first stage of $G$ to identify the target object/location in $I_r$. With the adverb $\mathcal{T}_{loc}$ related to locations extracted from $\mathcal{T}$ via CoreNLP~\cite{manning2014stanford}, it can be further encoded as $f^\mathcal{T}_{loc}$, describing the embedding of the location of interest with a pre-trained BERT~\cite{devlin2018bert}. Together with $I_r$, $L$ learns to mask out the object of interest by producing a binary mask $M = L(I_r, \mathcal{T}_{loc})$. More precisely, this is achieved by having $L$ perform cross-modal attention between $f^\mathcal{T}_{loc}$ and the feature map of $I_r$, followed by a mask decoder to produce $M$.

Unfortunately, it would be difficult to verify the correctness of the aforementioned mask without the presence of the target image. During the training of our cManiGAN, we apply a standard classification objective $\mathcal{L}^L_{in} = \mathcal{L}_{\text{CE}}(\text{MLP}(E(M\odot I_r)), y^r_{in})$ for the masked part, and the multi-label classification loss $\mathcal{L}^L_{out} = \mathcal{L}_{\text{BCE}}(\text{MLP}(E((1-M) \odot I_r), y_{out})$ for the unmasked region. We have $E$ as the feature extractor (e.g., VGG~\cite{simonyan2014very}), $y^r_{in}$ and $y_{out}$ denote the one/multi-hot label vectors indicating object category/categories in the masked/unmasked parts in $I_r$ and $I_t$, respectively. In the above derivation, MLP denotes multi-layer perceptron, and $\odot$ indicates element-wise dot product. And, $\mathcal{L}_{\text{CE}}$ and $\mathcal{L}_{\text{BCE}}$ represent the cross-entropy/binary-cross-entropy losses. Thus, the objective for learning the localizer $L$ is calculated by summing up $\mathcal{L}^L_{in}$, and $\mathcal{L}^L_{out}$. Thus, with the design and deployment of the Localizer, we are able to enforce the generator to manipulate the location of interest only. As later verified in our ablation studies, this allows our model to improve the generating quality and the feasibility of image manipulation under a weakly supervised setting.

\subsubsection{Image Inpainter $P$.} As the second stage in $G$, we have an image inpainter $P$ which takes the text feature $f^\mathcal{T}_{how}$ extracted from $\mathcal{T}$ by pre-trained BERT and the masked input $(1-M)\odot I_r$ for producing $I_g$. In addition to the standard GAN loss~\cite{goodfellow2014generative} for the generator $G$ with a discriminator $D$ deployed, we also calculate the reconstruction loss (i.e., mean squared error) $\mathcal{L}^P_{rec}=\mathcal{L}_{\text{MSE}}((1-M)\odot I_r, (1-M)\odot I_g)$, preserving the content of unmasked image regions $(1-M)\odot I_g$. Moreover, with an auxiliary classifier $\mathcal{C}$  jointly trained with the discriminator, we calculate the following classification losses: $\mathcal{L}^P_{out}=\mathcal{L}_{\text{BCE}}(\mathcal{C}((1-M) \odot I_g), y_{out})$, ensuring the semantic correctness of the unmasked image regions, and  $\mathcal{L}^P_{in}=\mathcal{L}_{\text{CE}}(\mathcal{C}((M\odot I_g), y^g_{in})$ to enforce that of the \textit{manipulated} object in $M\cdot I_g$. Note that $y^g_{in}$ denotes the one-hot label vector indicating the ground truth category of the object within the target location in $I_g$. 

With the above design, we have the total objective for $P$ as the sum over $\mathcal{L}_{GAN}, \mathcal{L}^P_{rec}, \mathcal{L}^P_{in}$, and $\mathcal{L}^P_{out}$, with both image global and local authenticities enforced by $D$~\cite{iizuka2017globally}. Please refer to the supplementary materials for the full objectives and implementation details.

\subsection{Cross-Modal Interpreter $I$}
\label{sec:interpreter}
For training target-free image manupulation models, how to preserve both visual and semantic correctness (e.g., visual attributes and spatial relationship) of the output image would be a challenging task. Thus, in addition to the above $G$ and $D$ modules, we introduce a cross-modal interpreter $\mathcal{I}$ in our cManiGAN for achieving this goal with additional word-level correctness enforced.\\

\noindent \textbf{Learning from Factual/counterfactual Descriptions.}
As shown in Fig.~\ref{Architecture_2}(b), given an input image, our cross-modal interpreter $\mathcal{I}$ learns to discriminate the factual description among multiple synthesized description candidates. This would enforce the manipulated image to be with sufficient visual and semantic correctness. To generate factual and counterfactual descriptions of an image, we utilize $\mathcal{T}$ and the labels for reference and target images (i.e., $O_r = \{o_1, o_2, ..., o_n \}$ and $O_t = \{o_1, o_2, ..., o_m \}$). We note that, with $n$ and $m$ represent the associated total numbers of objects, we have $|m-n| \leq 1$ since each instruction is considered to manipulate a single object. Thus, we define a basic description template as follows: ``There is a [OBJ] [LOC]", where [OBJ] can be replaced by the object label, and [LOC] indicates the adverbs of place to describe where [OBJ] is. 

With the above definitions, we synthesize the factual description $\mathcal{S}^f$ by replacing the [LOC] with adverbs of the place of the given instruction, extracted by CoreNLP~\cite{manning2014stanford}. We then replace the [OBJ] with the objects $o^f_r$ and $o^f_t$ for the reference and target images to generate the corresponding descriptions separately. The $o^f_r$ and $o^f_t$ can be identified by comparing the difference between $O_r$ and $O_t$ with the following three principles: 
\begin{itemize}
\item if $n > m$, $o^f_r$ = $O_r - O_t$ and $o^f_t$ is NONE.
\item if $n = m$, $o^f_r = O_r - (O_r \cap O_t)$ and $o^f_t = O_t - (O_r \cap O_t)$.
\item if $n < m$, $o^f_r$ is NONE, and $o^f_t = O_t - O_r$.
\end{itemize}
Note that NONE denotes the dummy category, which implies no object of interest at that location. 

As for synthesizing the counterfactual descriptions, we collect object tokens (e.g., green sphere and red cube in the CLEVR dataset) and relation tokens (e.g., in front of and behind) by applying NLTK tools~\cite{bird2009natural} on the factual ones. With these tokens, each object/relation counterfactual descriptions $\mathcal{S}^c_i$ ($i$ denotes the counterfactual description index) can be generated by randomly replacing the existing object/relation tokens from the factual descriptions with other non-existing tokens. As a result, a set of counterfactual descriptions $\mathbb{S}^c=\{\mathcal{S}^c_1, \mathcal{S}^c_2, ..., \mathcal{S}^c_N\}$ can be obtained by repeating the above process. Note that $N$ is the total number of counterfactual descriptions.\\

\begin{figure*}[t]
  \centering
  \includegraphics[page=1,trim={80 35 80 55}, clip, width=0.95\textwidth]{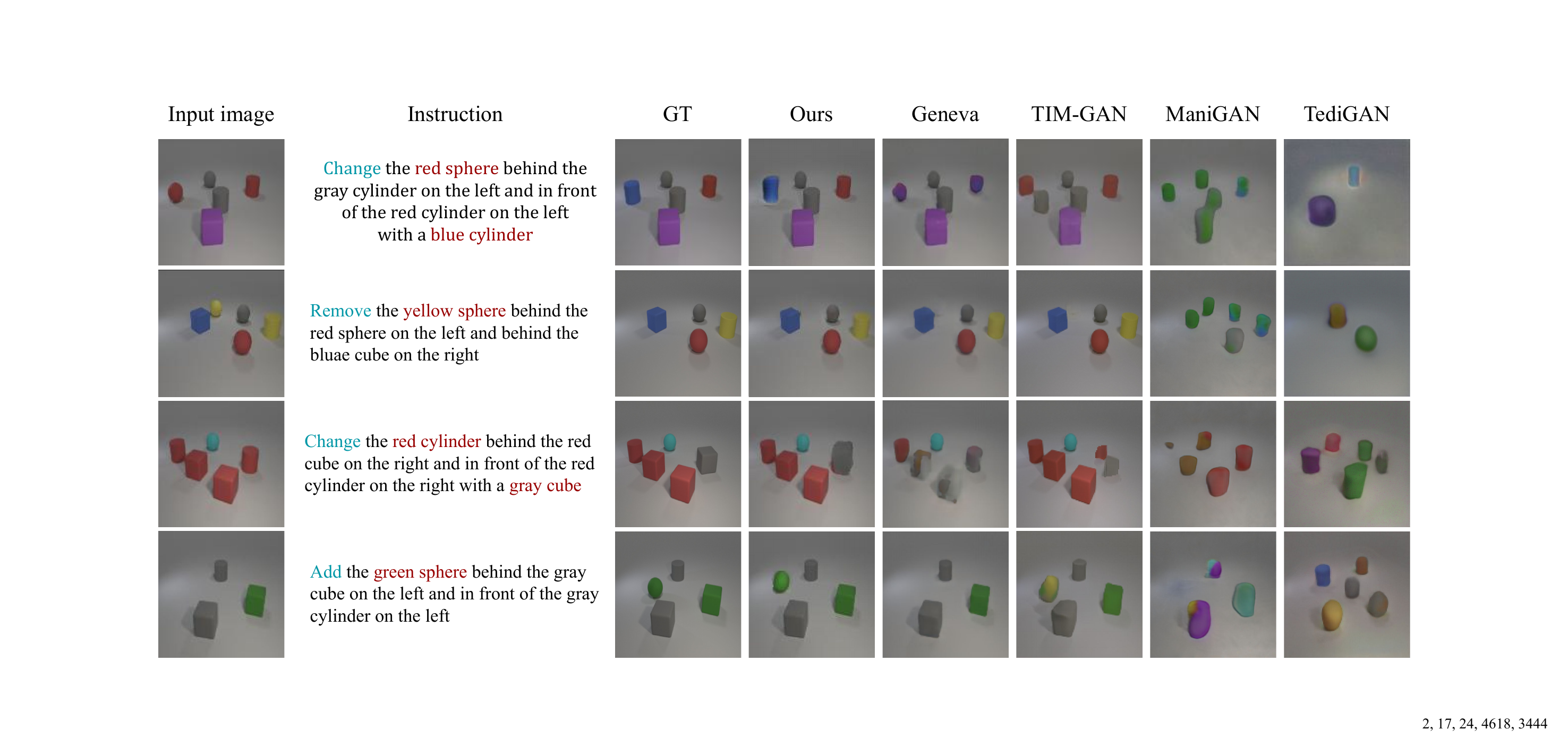}
  \vspace{-4mm}
  \caption{\textbf{Qualitative evaluation on the CLEVR dataset.} Each row shows the input reference image, instruction, ground truth (target) image and those generated by different methods. Note that GeNeVa~\cite{el2019tell} and TIM-GAN~\cite{zhang2021text} require target images during training. And, ManiGAN~\cite{li2020manigan} and TediGAN~\cite{xia2021tedigan} are mainly designed to tackle object-centric image data.}
  \vspace{-5mm}
  \label{exp:QL_CLEVR}
\end{figure*}

\noindent \textbf{Authenticating Semantic Correctness of $I_g$.} With the interpreter $\mathcal{I}$ taking the generated image $I_g$ and a set of descriptions $\mathbb{S}=\{\mathcal{S}^f, \mathcal{S}^c_1, \mathcal{S}^c_2, ..., \mathcal{S}^c_N\}$ (i.e., one factual description and $N$ counterfactual ones) as the inputs, our cManiGAN is able to assess the semantic correctness of $I_g$ by calculating the cross-modal matching scores $\hat{y}$ between the generated image $I_g$ and each of the descriptions in $\mathbb{S}$. For the architecture of $\mathcal{I}$, it can be viewed as a cross-modal alignment module $\Gamma$ (e.g., ViLBERT~\cite{lu2019vilbert} or a word-level discriminator in LWGAN~\cite{li2020lightweight}), which takes an image and language description/caption as inputs for producing the associated matching scores. Thus, the output scores $\hat{y}$ can be calculated as:

\begin{equation}
    \begin{aligned}
        \hat{y} = \mathcal{I}(I, \mathbb{S})
        = [  \Gamma(I,\mathcal{S}^f ), \Gamma(I,\mathcal{S}^c_1 ), \Gamma(I,\mathcal{S}^c_2 ), 
        ..., 
        \Gamma(I,\mathcal{S}^c_N )]. 
    \end{aligned}
\end{equation}

To train our two-stage editor with the interpreter $\mathcal{I}$, we take $\mathcal{I}$ as an auxiliary classifier along with $D$. Similar to ACGAN~\cite{odena2017conditional}, the loss function $\mathcal{L}_\mathcal{I}$ of the interpreter can thus be formulated as follows:
\begin{equation}
    \begin{aligned}
        \mathcal{L}_\mathcal{I} 
        =  \mathcal{L}_{\text{CE}}(\mathcal{I}(
        &I_r, \mathbb{S}_r),  y) + \mathcal{L}_{\text{CE}}(\mathcal{I}(I_g, \mathbb{S}_t), y).
    \end{aligned}
\end{equation}
 Note that $\mathbb{S}_r$ and $\mathbb{S}_t$ denote the synthesized description sets for reference and target images, respectively. Also, $y$ is one-hot vector with the only nonzero entry associated with the factual description.

\subsection{Reasoner $R$}
\textbf{Operational Cycle Consistency.}
With the deployment of the generator $G$, discriminator $D$, and the interpreter $I$, our cManiGAN is able to produce images with semantic authenticity preserved at the \textit{image} level. To further enforece the correctness at the \textit{pixel} level, we further introduce a reasoner $R$ in cManiGAN. As depicted in Fig.~\ref{Architecture_2}(a), this reasoner $R$ is designed to predict the undo instruction $\mathcal{T}'$ from $I_r$ and $\mathcal{T}$ (together with $O_r$ and $O_t$), which outputs the reconstructed version $I_{rec}$ and 
observe operational cycle consistency by minimizing the difference between $I_r$ and $I_{rec}$. Thus, this consistency objective $\mathcal{L}_{cyc}$ is calculated as:
\begin{equation}
    \begin{aligned}
        \mathcal{L}_{cyc} 
        = \mathcal{L}_{MSE}(I_r, I_{rec}) + \mathcal{L}_{perc.}(I_r, I_{rec}),
    \end{aligned}
\end{equation}
where $I_{rec} = G(G(I_r, \mathcal{T}), \mathcal{T}')$. Also, $\mathcal{L}_{MSE}$ and $\mathcal{L}_{perc.}$ represent the mean squared error and the perceptual loss~\cite{johnson2016perceptual}, respectively.\\

\noindent \textbf{Learning from Sequence-to-sequence Models.} Since the undo instruction is a textual sequence, we approach this reasoning task by solving a sequence-to-sequence learning problem and adopt the recent model of T5~\cite{raffel2019exploring} as the base model. However, since sequence models like T5 are trained on the language crawled corpus~\cite{raffel2019exploring}, they cannot be directly applied for text-guided image manipulation. Thus, as shown in Fig.~\ref{Architecture_2}(c), we design two learning tasks which adapt T5 for our reasoning task. First, we consider the \textit{image-level labels prediction} task, which predicts the labels $O_t$ by observing $O_r$ and $\mathcal{T}$, aiming at relating the changes in image-level labels. To equip $R$ with the ability to express its observation in terms of desirable instructions, we consider \textit{instruction generation} as the second fine-tuning task, with the goal to synthesize the full instruction by inputting $O_r$, $O_t$ and the adverbs of place part of instruction $\mathcal{T}_{loc}$ (extracted by CoreNLP~\cite{manning2014stanford} as noted earlier). With these two fine-tuning tasks, our reasoner $R$ is capable of inferring the undo instruction $\mathcal{T}'$ by observing $O_t$, $O_r$ and $\mathcal{T}$.

We now detail the learning process for our reasoner $R$. As shown in Fig.~\ref{Architecture_2}(c), to realize the sequence-to-sequence training scheme, we first convert image labels into pure text format with the subject-verb-object (SVO) sentence structure. Take $O_r=\{$purple sphere, green cube$\}$ for example, we have the text format as ``The labels for reference image contains purple sphere, green cube." Also, we denote the $\mathcal{T}^O_r$ and $\mathcal{T}^O_t$ as the text format of $O_r$ and $O_t$, respectively. With the labels desribed in text format, we construct the input text sequences $\hat{\mathcal{T}}$ for the fine-tuning tasks. For \textit{image-level prediction}, we have $\hat{\mathcal{T}}=\mathcal{T}^O_r \oplus \mathcal{T}$ serve as the input context, the T5 model is learned to predict $\mathcal{T}^O_t$. As for the second task of \textit{instruction generation}, the input context would be $\hat{\mathcal{T}}=\mathcal{T}^O_r \oplus \mathcal{T}^O_t \oplus \mathcal{T}_{loc}$, and the output would be the given instruction $\mathcal{T}$, where $\oplus$ is the concatenation on text. Note that the input text sequence for each fine-tuning task can be further created by combining a task-specific (text) prefix (e.g., what is the instruction) with the context according to the task of interest. Please refer to the supplementary materials for details and more training examples.

With the above designs, we impose the conventional sequence-to-sequence objective $L_{s2s}$~\cite{raffel2019exploring} to fine-tune the T5 model. Thus, the objective for learning $R$ can be formulated as follows:
\begin{equation}
    \begin{aligned}
        \mathcal{L}_{R} 
         = 
        \mathcal{L}_{s2s}(R(\mathcal{T}^O_r \oplus & \mathcal{T}^O_t \oplus \mathcal{T}_{loc}),  \mathcal{T}) \\
        &+ \mathcal{L}_{s2s}(R(\mathcal{T}^O_r \oplus \mathcal{T}), \mathcal{T}^O_t).
    \end{aligned}
\end{equation}

\begin{table*}[t]
\begin{center}
\vspace{0mm}
\resizebox{0.97\textwidth}{!}{
\setlength{\tabcolsep}{0.1mm}{
\begin{tabular}{l|ccccccc|ccccccc}
Operation  & \multicolumn{7}{c|}{Type 1: remove + add} & \multicolumn{7}{c}{Type 2: attribute change / shape} \\ \hline
Matrics      & FID  ↓         & \multicolumn{1}{c|}{IS ↑}    & \begin{tabular}[c]{@{}c@{}}image \\ acc (\%)\end{tabular} & \begin{tabular}[c]{@{}c@{}}In-mask \\ acc (\%)\end{tabular} & \multicolumn{1}{c|}{\begin{tabular}[c]{@{}c@{}}Interp. \\ acc (\%)\end{tabular}} & \multicolumn{1}{c}{R@1} & \multicolumn{1}{c|}{R@5} & FID  ↓         & \multicolumn{1}{c|}{IS ↑}    & \begin{tabular}[c]{@{}c@{}}image\\ acc (\%)\end{tabular} & \begin{tabular}[c]{@{}c@{}}In-mask\\ acc (\%)\end{tabular} & \multicolumn{1}{c|}{\begin{tabular}[c]{@{}c@{}}Interp.\\ acc (\%)\end{tabular}} & R@1 & R@5 \\[1ex] \hline 
Upper bound  & -              & \multicolumn{1}{c|}{-} & 99.25 & 88.66 & \multicolumn{1}{c|}{67.16} & \multicolumn{1}{c}{72.12}    & \multicolumn{1}{c|}{99.77}    & -              & \multicolumn{1}{c|}{-}              & 98.71 & 90.91 & \multicolumn{1}{c|}{67.19} & 96.27 & 99.85 \\ \hline
GeNeVa$^\dagger$     & {54.80}          & \multicolumn{1}{c|}{2.336} & {\ul92.93} & {\ul40.08} & \multicolumn{1}{c|}{34.27} & 33.32 & 79.23 & {\ul52.91} & \multicolumn{1}{c|}{2.017}          & 88.65 & {\ul7.18} & \multicolumn{1}{c|}{{\ul11.18}} & {\ul64.17} & {\ul76.75} \\
TIM-GAN$^\dagger$ & \textbf{43.38} & \multicolumn{1}{c|}{2.192} & 93.40 & 25.50 & \multicolumn{1}{c|}{\ul{38.17}} & {\ul33.72} & {\ul80.81} & {54.66} & \multicolumn{1}{c|}{2.122} & {\ul90.05} & 4.67 & \multicolumn{1}{c|}{10.79} & 58.73 & 76.37 \\
ManiGAN & 168.5 & \multicolumn{1}{c|}{\ul2.390} & 75.68 & 20.12 & \multicolumn{1}{c|}{0.88} & 0.01 & 0.09 & 170.1 & \multicolumn{1}{c|}{\ul2.234} & 73.78 & 2.3 & \multicolumn{1}{c|}{0.42} & 0.08 & 0.17 \\
TediGAN & 172.2 & \multicolumn{1}{c|}{\textbf{2.760}} & 69.60 & 26.07 & \multicolumn{1}{c|}{4.02} & 0.01 & 0.49 & 168.1 & \multicolumn{1}{c|}{\textbf{2.672}} & 69.47 & 2.46 & \multicolumn{1}{c|}{0.76} & 0.04 & 0.64 \\ 
\hline
Ours & {\ul45.88}  & \multicolumn{1}{c|}{2.214} & \textbf{93.59} & \textbf{43.01} & \multicolumn{1}{c|}{\textbf{40.85}} & \textbf{47.95} & \textbf{94.04} &  \textbf{38.26} & \multicolumn{1}{c|}{2.210} & \textbf{93.18} & {\textbf{39.18}} & \multicolumn{1}{c|}{{\textbf{33.74}}} & \textbf{87.46} & \textbf{94.01}  
\end{tabular}}}
\vspace{-2mm}
\caption{\textbf{Quantitative evaluation on CLEVR.} Note that \textbf{R@N} indicates the recall of the true target image in the top-N retrieved images. $^\dagger$ denotes methods requiring target images for training. The numbers in bold indicate the best scores, and those with underlines denote the second highest ones.}
\label{exp:QT_CLEVR}
\vspace{-6mm}
\end{center}
\end{table*}

\begin{table}[]
\begin{center}
\setlength{\tabcolsep}{0.7mm}{
\begin{tabular}{l|cc|ccc}
            & FID  ↓ & IS ↑ & \begin{tabular}[c]{@{}c@{}}image \\ acc (\%)\end{tabular} & \begin{tabular}[c]{@{}c@{}}Inside-mask\\ acc (\%)\end{tabular} & \begin{tabular}[c]{@{}c@{}}Inpterp.\\ acc (\%)\end{tabular} \\ \hline
\begin{tabular}[c]{@{}c@{}}Upper\\ bound\end{tabular} & - & - & 91.47 & 92.49 & 68.71 \\ 
\hline
Ours        & 166.18 & 4.64 & 86.04 & 17.17 & 13.54 \\
\hline
ASE$^\dagger$         & 132.04 & 6.37 & 86.99 & 41.66 & 33.34 \\
Ours$^\dagger$        & \textbf{104.77} & \textbf{7.21} & \textbf{89.73} & \textbf{50.03} & \textbf{46.20} 
 
\end{tabular}}
\vspace{-2 mm}
\caption{\textbf{Quantitative results on COCO.} $^\dagger$ denotes only the ``remove" operation is considered during evaluation.}
\vspace{-6 mm}
\label{exp:QT_COCO}
\end{center}
\end{table}

\noindent Once the above model is fine-tuned as our reasoner $R$, the undo instruction $\mathcal{T}'$ can be directly inferred by observing input test sequence as $\hat{\mathcal{T}}=\mathcal{T}_{loc} \oplus \mathcal{T}^O_r \oplus \mathcal{T}^O_t$ with the labels for the reference and target images swapped. 

As illustrated in Fig.~\ref{Architecture_2}(a), operational cycle consistency can be observed during the training of cManiGAN, providing additional desirable pixel-level guidance. For complete learning details (including pseudo code) of our cManiGAN, please refer to the supplementary material.

\section{Experiment}
\label{sec:exp}

\subsection{Datasets}

\subsubsection{CLEVR} The CLEVR dataset~\cite{johnson2017clevr} is created for multimodal learning tasks such as visual question answering, cross-modal retrieval, and iterative story generation. We consider the synthesized version of CLEVR as TIM-GAN~\cite{zhang2021text} did, which contains a total of 24 object categories (red cube, blue sphere, and cyan cylinder, etc.) with about 28.1K/4.6K paired synthesized images for training/validation.  Each training sample includes two paired images (reference image and target image (for evaluation only)) and an instruction describing where and how to manipulate the reference image.

\subsubsection{COCO}
The COCO dataset~\cite{lin2014microsoft} contains 118k real-world scene images for training with a total of 80 thing categories (car, dog, etc.). For simplicity, we consider a sampled COCO dataset containing 20 object categories (overlapped with Pascal-VOC~\cite{everingham2015pascal}) with about 12K/3K samples for training/validation. 
Since the target images are not available for COCO, only a reference image and an instruction are included in a training/validation sample (see supplementary material for the details). 

Note that three types of manipulations/operations, i.e., “add”, “remove”, and “change”, are considered in both datasets. We will make the datasets publicly available for reproduction and comparison purposes. 

\subsection{Qualitative Evaluation}
We compare our proposed cManiGAN with recent models, including GeNeVa~\cite{el2019tell}, TIM-GAN~\cite{zhang2021text}, ManiGAN~\cite{li2020manigan} and TediGAN~\cite{xia2021tedigan}, with example results shown in Fig.~\ref{exp:QL_CLEVR}. From this figure, we observe that the outputs of the attribute-based methods, such as ManiGAN and TediGAN, were not able to locate proper image regions for manipulation and even failed to preserve the image structure or details. As for the structure-based methods (i.e., GeNeVa and TIM-GAN), even though the structure of the reference image was preserved, these methods fail to comprehend the complex input instruction and lack the ability to manipulate the image with visual and semantic correctness. Take the third case in Fig.~\ref{exp:QL_CLEVR} for example, GeNeVa incorrectly changed the visual attributes of two non-target objects. On the other hand, our cManiGAN was able to generate consistent outputs following the given instructions. For more qualitative results, please refer to our supplementary materials.

\subsection{Quantitative Evaluation}
To quantify and compare the performances between different models, two metrics are applied: Fr\'{e}chet inception distance (FID) and inception score (IS). Moreover, the following four different metrics are utilized for evaluation:\\ 
(1) \textbf{Image classification accuracy (Image acc)} measures whether the objects in the generated image match the labels of the target image. (2) \textbf{Inside mask classification accuracy. (In-mask acc)} evaluates whether the generated object in the masked part can be recognized by a pre-trained classification model. (3) \textbf{Interpreter accuracy (Interp. acc).} measures whether the generated image semantically matches its factual description via a cross-modal interpreter, which is pre-trained on the reference image $I_r$ and its corresponding description set $\mathbb{S}_r$. (4) \textbf{Retrieval score (RS)} evaluates the manipulation correctness of the manipulation by applying the existing text-guided image retrieval method of TIRG~\cite{vo2019composing}. Following TIM-GAN~\cite{zhang2021text}, we report RS@N, where N indicates the recall of the ground-truth image in the top-N retrieved images.

\subsubsection{Quantitative Comparisons Image Quality and Realness.}
In Table~\ref{exp:QT_CLEVR}, we compare our cManiGAN with GeNeVa~\cite{el2019tell}, TIM-GAN~\cite{zhang2021text}, ManiGAN~\cite{li2020manigan} and TediGAN~\cite{xia2021tedigan} in terms of FID, IS, image accuracy, and inside-mask accuracy. To better compare these methods for image manipulation, we consider structure-based and attribute-based operations (as mentioned in Sect.~\ref{sec:intro}), and we divide the evaluation into two types: Type 1 focuses on the ``remove/add" operations, and Type 2 considers only ``attribute/shape change" operations. These metrics measure the quality of the generated results, reflecting the resulting realness of the synthesized images. From this table, we see that our cManiGAN reported comparable or improved FID and IS scores on both types of operations. It is worth pointing out that, while the best FID scores were reported by TIM-GAN~\cite{zhang2021text} in Type 1 comparisons, they require \textit{ground-truth target images} during training, while others (including ours) do not have such a requirement.

To further quantify the output accuracy, we report the multi-label classification accuracy on the produced images, and also the single-label classification accuracy on the image regions masked by the ground truth masks. As listed in the above table, we see that cManiGAN achieved either the best or the second-best image accuracy and in-mask accuracy among all methods on two types (i.e, Type 1 as add/remove and Type 2 as attribute/shape change) of operations, which verify that our method successfully edits the image with visual correctness. We observe that GeNeVa tended to ``remove" objects in the reference image not following the instruction, resulting in degraded image-level accuracy even with satisfactory in-mask accuracy on Type 1 operations (i.e, add/remove). This is mainly due to the fact that it is optimized on such operations and cannot easily generalized to complex ones like ``add".

\subsubsection{Semantic Relevance and Correctness.}
In Table~\ref{exp:QT_CLEVR}, we report the accuracy of the interpreter and retrieval scores to measure the semantic relevance between the manipulated images and the instructions, as well as the visual correctness of the generated objects. From this table, we see that our cManiGAN again outperformed favorably against baseline methods across both types of operations (i.e., add/remove and attribute/shape change). While GeNeVa and TIM-GAN produced visually realistic images, their corresponding interpretation accuracy and retrieval scores were not satisfactory, indicating their lack of ability in comprehending complex instructions to edit the images with semantic and visual correctness. In comparison, our approach consistently achieved higher scores on interpretation accuracy and retrieval scores with remarkable margins. Additional user studies in terms of generating realism and semantic relevance are presented in the supplementary materials.

\subsection{Real-World Target-Free Images}
We now consider the COCO dataset, a challenging real-world dataset with no target image data (i.e., no retrieval scores R@N can be measured). For quantitative comparison purposes, we compare with ASE~\cite{shetty2018adversarial}, which focuses on object removal in the scene image from a weakly supervised setting, and the results are shown in Table~\ref{exp:QT_COCO_ablation}(a). From this table, we see that our cManiGAN outperformed ASE in terms of both image quality and semantic correctness. More specifically, our method improved the FID score while the interpretation accuracy was improved by nearly 10\%.

\begin{figure}[t]
  \centering
  \includegraphics[page=2, trim={180 60 180 60}, clip, width=0.45\textwidth]{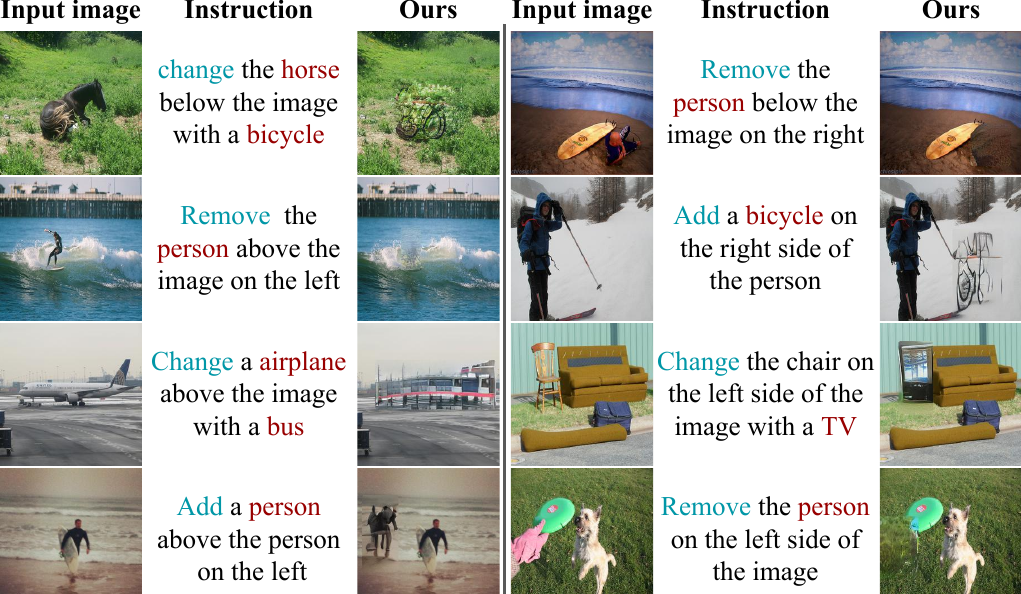}
  \vspace{-2mm}
    \caption{Qualitative examples on the COCO dataset.}
    \vspace{-3mm}
    \label{exp:QL_COCO}
\end{figure}

As for qualitative analysis, we provide example results of our cManiGAN in Fig.~\ref{exp:QL_COCO} (more quantitative and qualitative results are available in the supplementary). From this figure, one can see that the manipulated scene images were with sufficient semantic and visual correctness according to the instructions. To the best of our knowledge, no existing works tackle image manipulation by text instruction on such real-world scene images. Also, please refer to supplementary materials for more experiments on COCO.

\begin{table}[!tb]
\vspace{2mm}
\centering
\begin{small}
\resizebox{0.45\textwidth}{!}{
\setlength{\tabcolsep}{0.2mm}{
\begin{tabular}{l|cc|ccc}
 & FID  ↓        & IS ↑    & \begin{tabular}[c]{@{}c@{}}image\\ acc (\%)\end{tabular} & \begin{tabular}[c]{@{}c@{}}Inside-mask\\  acc (\%)\end{tabular} & \begin{tabular}[c]{@{}c@{}}Interp. \\ acc (\%)\end{tabular} \\ \hline
Upper bound & - & - & 98.96 & 89.56 & 67.16 \\ \hline 
Ours w/o $L$ & 228.7 & 1.11 & 72.52 & 24.38 & 0.667 \\
Ours w/o R (cycle) & 68.08 & 2.07 & 83.47 & 41.40 & 28.27 \\
Ours w/o $\mathcal{I}$ & 44.08 & 2.11 & \textbf{93.73} & 41.01 & 35.14 \\
Ours w/o R, $\mathcal{I}$ & 77.56 & 2.08 & 80.22 & 39.70 & 26.55 \\
\textbf{Ours} & \textbf{39.41} & \textbf{2.22} & 93.41 & \textbf{41.92} & \textbf{37.46} \\
\end{tabular}}}
\end{small}
\vspace{-2mm}
\caption{\textbf{Ablation studies on CLEVR}. Note that $L$, $\mathcal{I}$ and cycle denote the localizer in generator, cross-modal interpreter and operational cycle-consistency losses, respectively.}
\label{exp:QT_COCO_ablation}
\vspace{-3mm}
\end{table}

\subsection{Ablation Studies and Remarks}

\subsubsection{Design of cManiGAN.}
Table~\ref{exp:QT_COCO_ablation}(b) assesses the contributions of each deployed module in our cManiGAN, and thus verifies the design of our proposed model. To justify the two stage design of our generator $G$, we ablate the localizer $L$, and utilize only the in-painter $P$ to produce the generated image by observing the reference image and the instruction. One can find that, by adding the localizer, our generator is able to comprehend the complex instruction and locate the target location, which enforces the generator to manipulate within target location only, improving both the semantic correctness of the generated image and the overall visual quality of the generated image. To verify the effectiveness of the interpreter $\mathcal{I}$ during training, we show that adding such a module (comparing the third and fifth rows of this table) into our baseline would further improve the performance, with improved visual realness and semantic correctness.

Finally, we verify the contribution of the operational cycle-consistency and observe improved visual quality scores in FID. This verifies that our operational cycle-consistency serves as a potential pixel-level training feedback without observing the target images. By combining all of the above designs, our full version of cManiGAN achieves the best results in Table~\ref{exp:QT_COCO_ablation}(b). Thus, the design of our cManiGAN can be successfully verified. For more ablation studies of objective function and reasoner module, please refer to our supplementary materials.

\section{Conclusion}

We presented a Cyclic Manipulation GAN (cManiGAN) for target-free text-guided image manipulation, realizing \textit{where} and \textit{how} to edit the input image with the given instruction. To address this task, a number of network modules are introduced in our cManiGAN. The image editor learns to manipulate an image in a two-stage manner, locating the image region of interest followed by completing the output image. In order to guarantee that the output image would exhibit visual and semantic correctness at pixel and world levels, our cManiGAN has unique modules of a cross-modal interpreter and a reasoner, leveraging auxiliary semantics self-supervision. The former associates the image and text modalities and serves as a semantic discriminator, which enforces the authenticity and correctness of the output image via word-level training feedback. The latter is designed to infer the ``undo" instruction, allowing us to train cManiGAN with operational cycle-consistency, providing additional pixel-level guidance. With extensive quantitative and qualitative experiments, including user studies, the use of the model is shown to perform favorably against state-of-the-art methods which require different degrees of supervision or be applicable to a limited amount of manipulation tasks. 

\clearpage

\section{Acknowledgement}

This work is supported in part by the National Science and Technology Council of Taiwan under grant NSTC 111-2634-F-002-020. We also thank the National Center for High-performance Computing (NCHC) and Taiwan Computing Cloud (TWCC) for providing computational and storage resources. Finally, we are especially grateful to Dr. Yun-hsuan Sung and Dr. Da-Cheng Juan from Google Research for giving their valuable research advice.

%
%
\bibliography{reference}

\end{document}